\documentclass[letterpaper, 10 pt, journal, twoside]{IEEEtran}

\usepackage{graphicx} 
\usepackage{float}  
\usepackage{subfigure}  

\graphicspath{{./images/}} 
\usepackage{multicol}
\usepackage{amsmath}
\usepackage{amsfonts}
\usepackage{soul}
\usepackage{hyperref}
\usepackage{makecell}

\usepackage{bm}

\usepackage{booktabs}
\usepackage{multirow}

\usepackage{tikz}

\newcommand\copyrighttext{%
	\footnotesize \textcopyright 2026 IEEE. Personal use of this material is permitted.
	Permission from IEEE must be obtained for all other uses, in any current or future
	media, including reprinting/republishing this material for advertising or promotional
	purposes, creating new collective works, for resale or redistribution to servers or
	lists, or reuse of any copyrighted component of this work in other works.}
\newcommand\copyrightnotice{%
	\begin{tikzpicture}[remember picture,overlay]
		\node[anchor=south,yshift=10pt] at (current page.south) 
		{\fbox{\parbox{\dimexpr\textwidth-\fboxsep-\fboxrule\relax}{\copyrighttext}}};
	\end{tikzpicture}%
}

\begin{document}
%
\title{Virtual-force Based Visual Servo for Multiple Peg-in-Hole Assembly with Tightly Coupled Multi-Manipulator}
%
%
%

\author{Jiawei Zhang$^{1}$, Chengchao Bai$^{1}$, Wei Pan$^{2}$, and Jifeng Guo$^{1}$%
\thanks{Manuscript received: August, 14, 2025; Revised: November, 20, 2025; Accepted: December, 16, 2025.}
\thanks{This paper was recommended for publication by Editor Markus Vincze upon evaluation of the Associate Editor and Reviewers' comments. This work was supported by (organizations/grants which supported the work.)} 
\thanks{$^{1}$Jiawei Zhang, Chengchao Bai, and Jifeng Guo are with Harbin Institute of Technology, China
        {\tt\footnotesize j.zhang@stu.hit.edu.cn}}%
\thanks{$^{2} $Wei Pan is with The Univeristy of Manchester, UK}%
\thanks{Digital Object Identifier (DOI): see top of this page.}
}

\markboth{IEEE Robotics and Automation Letters. Preprint Version. Accepted December, 2025}
{Zhang \MakeLowercase{\textit{et al.}}: VF-based Visual Servo for MPiH Assembly with Tightly Coupled Multi-Manipulator}

\maketitle
\copyrightnotice
\begin{abstract}
Multiple Peg-in-Hole (MPiH) assembly is one of the fundamental tasks in robotic assembly. In the MPiH tasks for large-size parts, it is challenging for a single manipulator to simultaneously align multiple distant pegs and holes, necessitating tightly coupled multi-manipulator systems. For such MPiH tasks using tightly coupled multiple manipulators, we propose a collaborative visual servo control framework that uses only the monocular in-hand cameras of each manipulator to reduce positioning errors. Initially, we train a state classification neural network and a positioning neural network. The former divides the states of the peg and hole in the image into three categories: obscured, separated, and overlapped, while the latter determines the position of the peg and hole in the image. Based on these findings, we propose a method to integrate the visual features of multiple manipulators using virtual forces, which can naturally combine with the cooperative controller of the multi-manipulator system. To generalize our approach to holes of different appearances, we varied the appearance of the holes during the dataset generation process. The results confirm that by considering the appearance of the holes, classification accuracy and positioning precision can be improved. Finally, the results show that our method achieves 100\% success rate in dual-manipulator dual peg-in-hole tasks with a clearance of 0.2 mm, while robust to camera calibration errors.
\end{abstract}


%

\section{Introduction}
\begin{figure}[t]
	\centering
	\setlength{\abovecaptionskip}{0.cm}
	\includegraphics[width=0.49\textwidth]{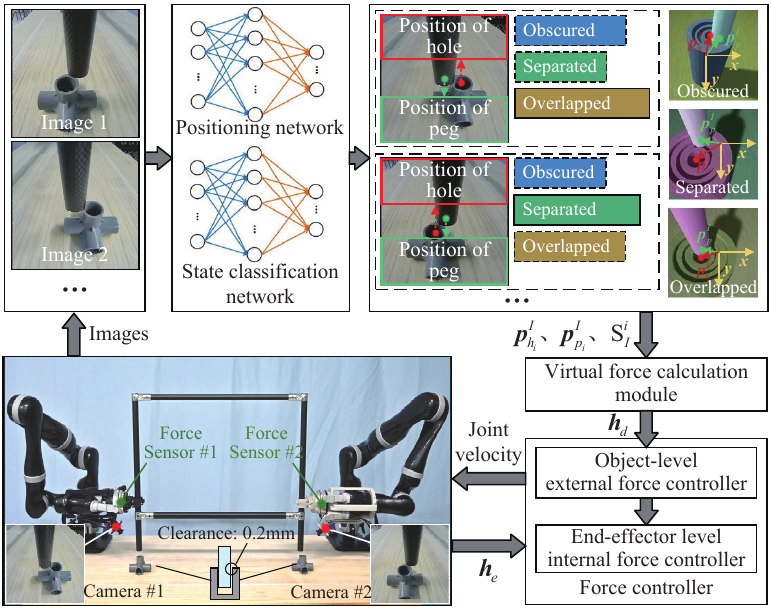} 
	\caption{The workflow of the proposed method.}
	\label{fig1}
	\vspace{-10pt}
\end{figure}

%
%
%
%
\IEEEPARstart{I}{n} recent years, robots have played an increasingly important role in automated assembly. The MPiH assembly is a fundamental operation in this context. Current research on MPiH typically focuses on assembling small parts using the single \cite{c1} or multiple manipulators \cite{c2}. The MPiH task for large-size parts is rarely studied, and a typical task is to assemble large-size trusses in orbit using space manipulators \cite{c3}. The main feature of the MPiH task of large-size parts is that multiple pegs and holes that are far apart need to be aligned simultaneously. If only a single manipulator is used, the heavy parts cannot be carried and the small movement of the manipulator may cause large displacement of the remote peg, which makes it difficult to meet the accuracy. When faced with such tasks, humans are accustomed to using both hands to grasp parts simultaneously for assembly. Inspired by this, this paper studies the use of tightly coupled multi-manipulators for assembly, as shown in Fig.~\ref{fig1}. In this paper, when all the end-effectors are fixedly connected to an object at the same time, these manipulators are said to be tightly coupled.

The MPiH tasks can be divided into two stages: hole searching and insertion. During the hole-searching stage, the positioning uncertainty should be reduced to align the pegs and holes. The insertion stage aims to complete the assembly compliantly while avoiding jamming. In this paper, we focus on the hole search stage. Compared with the Single Peg-in-Hole (SPiH) assembly task, the MPiH task of large size parts is more challenging. Firstly, there are complex contact states in the MPiH task \cite{c4}, and it is difficult to use contact force information to reduce positioning errors as in the SPiH task, and visual information is indispensable. It is necessary to study the accuracy and robustness of visual feature extraction. Secondly, when tightly coupled multiple manipulators are used for assembly, the images come from in-hand cameras of different manipulators. In this case, how to integrate visual features to efficiently guide the movement of the manipulators and quickly reduce positioning errors is a key problem. To solve the above problems, we study how to improve the accuracy and robustness of neural networks in MPiH task, and propose the concept of virtual force to integrate visual features at different positions. Specifically, the main contributions of this paper are as follows.

\begin{enumerate}
	
	\item We created a new synthetic dataset for the peg-in-hole tasks, enhancing previous datasets by introducing variations in hole appearances. Compared to training only on flat holes, our method has a better performance.
	
	\item For the MPiH tasks using tightly coupled multiple manipulators, a collaborative visual servo control framework is proposed based on virtual force. Which requires only a monocular in-hand camera and can naturally integrate the visual features of different manipulators, while being robust to camera calibration errors. This framework assumes that the pegs are fixed to the end-effectors and there is no relative motion between them.
	
\end{enumerate}

\section{RELATED WORKS}

\subsection{SPiH Assembly using Single Manipulator}

There are many methods of searching for holes for the SPiH tasks. Blind search using pre-designed search trajectories, such as spiral trajectories \cite{c5}, Lissajous curves \cite{c6}, or others, to explore areas where holes may exist. Such methods typically require a long time because of the absence of prior information about the hole position. The reinforcement learning based hole search methods typically take the contact force and estimated hole position as input \cite{c7}. They define the action space as a set containing a limited number of action primitives and require a large amount of interaction data for training. Recent research shows that finite-state machines can be used to select action primitives directly, which are simpler and more efficient \cite{c8}. Imitation learning is a technique that directly learns skills from demonstration data. Common methods include Dynamic Movement Primitives (DMPs) \cite{c9}, Gaussian Mixture Regression (GMR) \cite{c10}, deep learning \cite{c11}, and so on. Compared to reinforcement learning, imitation learning has higher data efficiency, but requires a lot of teaching data. ARIE \cite{c12} is a concept inspired by human assembly, which automatically reduces the pose error between the peg and the hole through environmental constraints. However, both the finite-state machine \cite{c8} and the ARIE can only be used when the end faces of the peg and hole overlap. 

Visual servo methods are capable of reducing the large pose error. Joshua et al. \cite{c13} used the VGG network to estimate the position of the hole and synthesized training images using domain randomization. They discretized the movement of the manipulator into finite motion directions to enhance the algorithm's robustness. In contrast, Rasmus et al. \cite{c14} simultaneously estimated the positions of both the peg and the hole. They aligned the peg and hole through a continuous visual servo, which proved to be more efficient.

\subsection{MPiH Assembly using Single Manipulator}

Research on the MPiH assembly using single manipulator has a long history. Sathirakul et al. \cite{c4} first studied the possible equilibrium states in the two-dimensional dual-peg insertion tasks, gave the geometric conditions and the force-torque equations of each equilibrium state, and obtained the jamming diagrams \cite{c15} and the taxonomy of dual-peg insertions. After that, Fei et al. \cite{c16} analyzed the three-dimensional multiple peg-in-hole assembly process, and Zhang et al. \cite{c17} analyzed the flexible dual peg-in-hole assembly process of flexible parts. The aforementioned studies focus on the assembly strategy based on the contact model. Due to the complexity of the contact model in the MPiH tasks, the assembly strategy not based on the contact model has received much attention in recent years \cite{c1,c18}. The aforementioned research typically focus on the insertion stage, which involves small-scale movements. Currently, there is limited literature on the hole search stage in the context of the MPiH tasks. Lee et al. \cite{c19} improved the spiral trajectory and proposed the Search Trajectory with a Twisting Motion (STTM) for the MPiH task. 

\subsection{SPiH Assembly using Multiple Manipulators}

In the SPiH tasks using multiple manipulators, one manipulator is typically designated as the assist arm, while the other serves as the task arm \cite{c5,c20,c21}. The assist arm is used to hold the part with the hole, whereas the task arm is used to hold the peg. In such assembly tasks, there are no closed-chain constraints between the assist arm and the task arm, resulting in a loosely coupled state. Lee et al. \cite{c5} used GMM to classify contact states in cooperative dual-arm assembly tasks, the joint torques of the assist arm and the task arm are considered during classification. Alles et al. \cite{c20} used model-free reinforcement learning to train a centralized policy in the simulation environment, which simultaneously controls the movements of the assist arm and the task arm, achieving direct transfer from simulation to reality. Consequently, Yao et al. \cite{c21} used a multi-agent reinforcement learning algorithm to train distributed policies for the assist arm and the task arm.

\subsection{MPiH Assembly using Multiple Manipulator}

Recently, there are also some studies on MPiH assembly using multiple manipulator. Based on the DMPs, Yang et al. \cite{c2} proposed a trajectory planning method for MPiH task of small parts using dual manipulators, and achieved a high success rate in the 2D plane. Ge et al. \cite{c22} used tightly coupled dual manipulators to assemble large parts, they focused on the controller design of the assembly process, and did not consider the influence of positioning error and the size of the parts on the assembly process. 

In summary, although there are many studies on peg-in-hole tasks, the MPiH tasks for large-sized parts using tightly coupled multi-manipulator is rarely studied. This paper focuses on the reduction of positioning errors in such tasks, which is an unstudied problem.

\section{METHODS}

\subsection{Overview of the Method}

The workflow of the proposed method is shown in Fig.~\ref{fig1}. It mainly consists of four parts: state classification neural network, positioning neural network, virtual force calculation module, and force controller for tightly coupled multi-manipulator. Firstly, the states of the peg and hole in the images are categorized into three classes: obscured, separated, and overlapped. Examples of the three types of states are shown in Fig.~\ref{fig1}.  The state classification neural network is used to determine the state of the peg and hole, and the positioning neural network is used to estimate the positions of the peg and hole in the images. Building upon this, we calculate the virtual force acting on the object, which is used as an input to the multi-manipulator cooperative controller to guide the object's motion. In the following, we will provide detailed explanations of these components.

\begin{figure}[t]
	\centering
	\setlength{\abovecaptionskip}{0.cm}
	\includegraphics[width=0.4\textwidth]{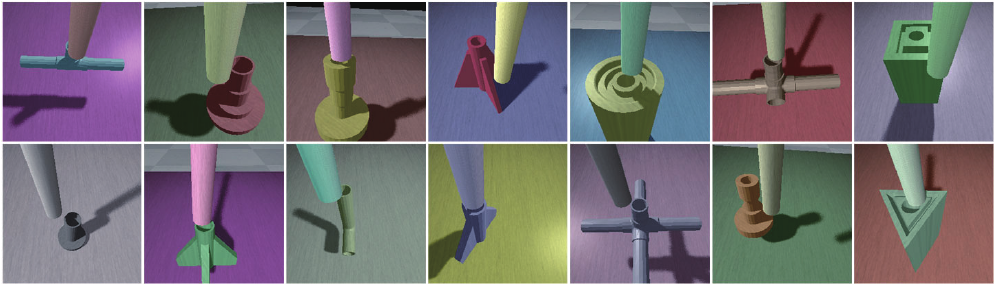} %
	\caption{The images of scenes with different hole appearances. From top left to bottom right: hole1 to hole14.}
	\label{fig2}
	\vspace{-10pt} %
\end{figure}

\begin{figure}[t]
	\centering
	\setlength{\abovecaptionskip}{0.cm}
	\includegraphics[width=0.4\textwidth]{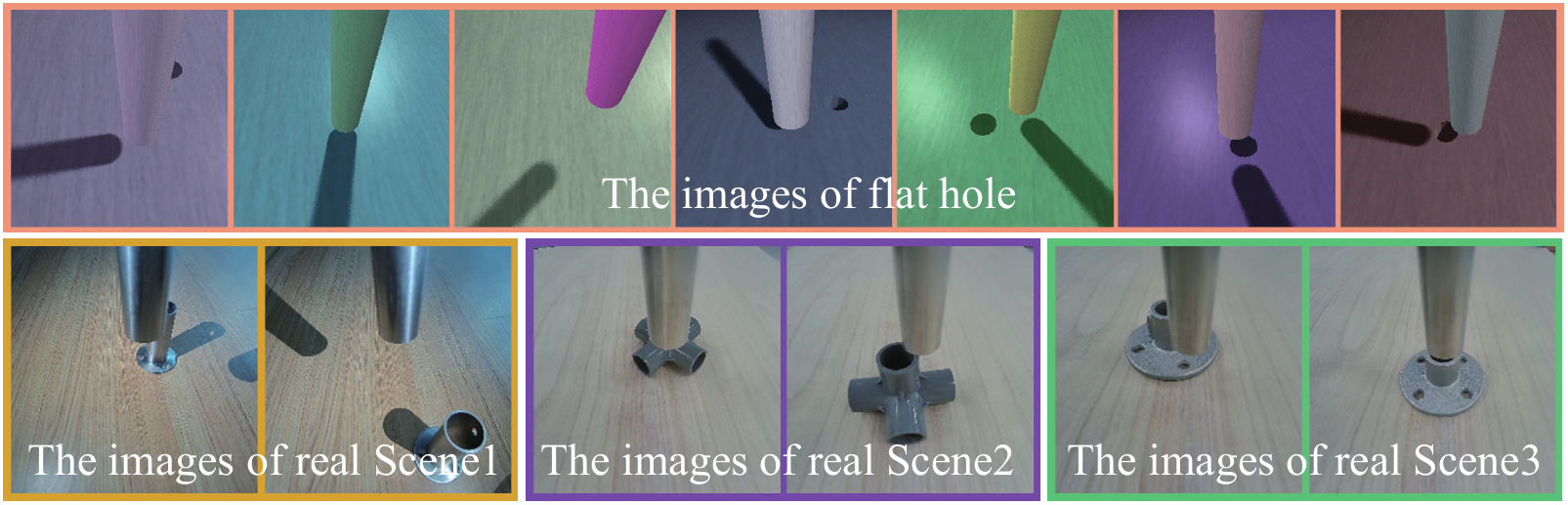} 
	\caption{Images of the flat hole scene and the real peg-in-hole scenes.}
	\label{fig3}
	\vspace{-15pt} 
\end{figure}

\subsection{The State Classification Neural Network and Positioning Neural Network}
\textit{1) Network structure:} For the state classification neural network, we use the ImageNet pretrained EfficientNetV2-S \cite{c23} . The neural network takes RGB images of size 224×224×3 as input, and the output of the neural network is changed from 1000 dimensions to 3 dimensions. We train the neural network using the cross-entropy loss function. For the positioning neural network, we use ResNetUNet \cite{c14} (a U-Net architecture with an ImageNet pretrained ResNet18 backbone). Similarly to \cite{c14}, we utilize the Gaussian kernel to convert the true positions of the peg and the hole into heatmaps to train the neural network. The positions of the peg and hole are represented as $\boldsymbol{p}_\mathrm{p}^I$ and $\boldsymbol{p}_\mathrm{h}^I$ in the pixel coordinate system, as shown in Fig.~\ref{fig1}. The positioning neural network takes RGB images of size 224×224×3 as input and outputs an estimated dual-layer heatmap $\hat{\boldsymbol{H}}$ of size 224×224×2. The Mean Squared Error (MSE) is chosen as the loss function. The positions of the maximum pixel values in each layer of the heatmap $\hat{\boldsymbol{H}}$ are taken as the estimated positions of the peg and the hole.

\textit{2) Data generation:} The neural network in this paper is trained entirely on images generated by the CoppeliaSim simulator with POV-Ray render mode. To mitigate the influence of hole appearances, we built 14 scenes with different hole appearances, along with a control scene with a flat hole. By randomizing the scene's color, light poses, camera poses, and peg poses, we generated 300 images for each hole appearance, comprising 100 images for each of the obscured, overlapped, and separated states. Some images of these 14 holes are shown in Fig.~\ref{fig2}. For the scene with the flat hole, we generated 1800 images, with 600 images allocated for each of the obscured, overlapped, and separated states. Some sample images of the flat hole are shown in Fig.~\ref{fig3}. In the CoppeliaSim, the position of the peg and hole in the world coordinate system is known. The $\boldsymbol{p}_\mathrm{p}^I$, $\boldsymbol{p}_\mathrm{h}^I$ and the relative state of the peg and hole can be automatically calculated using the imaging model of the camera. The code generating the dataset is available at: \url{https://github.com/hit618/peg-in-hole-images-generator.git}.

Additionally, we manually annotated images from three real peg-in-hole scenes, each containing 300 images, with 100 images each for obscured, separated, and overlapped states. We name the three real datasets as Real 1-3, and some images of them are shown in Fig.~\ref{fig3}.

\subsection{Virtual Force Calculation Module}

Suppose that the number of manipulators is $m$. the position of the pegs and holes in the pixel coordinate system is denoted as $\boldsymbol{p}_{p_i}^I$ and $\boldsymbol{p}_{h_i}^I (i=1,2,\cdots,m)$. The position vectors of the pegs and holes in the camera coordinate system are denoted as $\boldsymbol{p}_{p_i}^c$ and $\boldsymbol{p}_{h_i}^c$, they can be calculated using $\boldsymbol{p}_{p_i}^I$ , $\boldsymbol{p}_{h_i}^I$, and the camera's imaging model. Then, the unit vectors $\boldsymbol{n}_{p_i}$ and $\boldsymbol{n}_{h_i}\:$ are calculated in the world coordinate: 
\begin{equation}
	\left\{ \begin{array}{l}
		{\boldsymbol{n}_{{p_i}}} = {\boldsymbol{R}_{{c_i}}}\boldsymbol{p}_{{p_i}}^c/\left\| {{\boldsymbol{R}_{{c_i}}}\boldsymbol{p}_{{p_i}}^c} \right\|\\
		{\boldsymbol{n}_{{h_i}}} = {\boldsymbol{R}_{{c_i}}}\boldsymbol{p}_{{h_i}}^c/\left\| {{\boldsymbol{R}_{{c_i}}}\boldsymbol{p}_{{h_i}}^c} \right\|
	\end{array} \right.
	\label{eq:1}
\end{equation}

\begin{figure}[t]
	\centering
	\setlength{\abovecaptionskip}{0.cm}
	\includegraphics[width=0.4\textwidth]{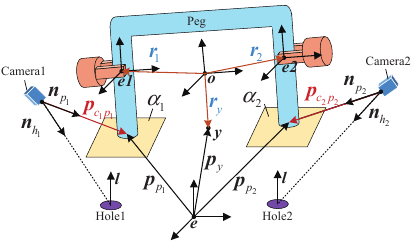} 
	\caption{Schematic diagram of the vectors in the paper. ${\Sigma _o}$ denotes the coordinate system of the center of mass of the object, ${\Sigma _e}$ denotes the world coordinate system. The superscript of each vector indicates the coordinate frame in which it is expressed, and if it is relative to the ${\Sigma _e}$, the superscript is omitted by default. }
	\label{fig4}
	\vspace{-15pt} 
\end{figure}

Where $\boldsymbol{R}_{c_i}$ is the rotation matrix representing the pose of the camera coordinate system relative to the world coordinate system, and the unit vector of the hole axis is represented as $\boldsymbol{l}.$ The position of the origin of the camera coordinate system in the world coordinate system is represented as $\boldsymbol{p}_{c_i}$. Suppose that the peg is fixed to the manipulator's end-effector and $\boldsymbol{p}_{{c_i}{p_i}}$ is the vector from camera position $\boldsymbol{p}_{c_i}$ to the peg position $\boldsymbol{p}_{p_i}$. The diagram of the vectors is illustrated in Fig.~\ref{fig4}. Let $\alpha_i$ be the plane that passes through $\boldsymbol{p}_{p_i}$ and perpendicular to $\boldsymbol{l}$. 
The intersection points of the vectors $\boldsymbol{n}_{p_i}$ and $\boldsymbol{n}_{h_i}$ with the plane $\alpha_i$ is represented as ${\boldsymbol{p}_{\alpha_i {p_i}}}$ and ${\boldsymbol{p}_{\alpha_i {h_i}}}$, respectively. ${\boldsymbol{p}_{{\alpha_i}{p_i}}}$ and ${\boldsymbol{p}_{\alpha_i {h_i}}}$ are calculated as follows:
\begin{equation}
	\left\{ \begin{array}{l}
		\boldsymbol{p}_{\alpha_i {p_i}} = {\boldsymbol{p}_{{c_i}}}{\rm{ + }}\frac{{{\boldsymbol{p}_{{c_i}{p_i}}} \cdot \boldsymbol{l}}}{{{\boldsymbol{n}_{{p_i}}} \cdot \boldsymbol{l}}}{\boldsymbol{n}_{{p_i}}}\\
		\boldsymbol{p}_{\alpha_i {h_i}} = {\boldsymbol{p}_{{c_i}}}{\rm{ + }}\frac{{{\boldsymbol{p}_{{c_i}{p_i}}} \cdot \boldsymbol{l}}}{{{\boldsymbol{n}_{{h_i}}} \cdot \boldsymbol{l}}}{\boldsymbol{n}_{{h_i}}}
	\end{array} \right.
	\label{eq:2}
\end{equation}

\begin{table}[t]
	\scriptsize
	\centering
	\setlength{\abovecaptionskip}{0.cm}
	\caption{The Parametes of The Controller}
	\label{table:1}
	\begin{tabular}{c c c c c c} 
		\toprule 
		Parameters & Value & Parameters & Value & Parameters & Value \\ 
		\midrule 
		${k_F}$ & 10 & $\boldsymbol{M}_d^y$ & ${\rm{10}}{\boldsymbol{I}_6}$ & $\boldsymbol{M}_d^i$ & ${\rm{50}}{\boldsymbol{I}_6}$\\
		${k_M}$ & 6 & $\boldsymbol{K}_d^y$ & ${\rm{1000}}{\boldsymbol{I}_6}$ & $\boldsymbol{K}_d^i$ & ${\rm{100}}{\boldsymbol{I}_6}$\\
		$F_c$ & 8 & $\boldsymbol{K}_p^y$ & ${\rm{10}}{\boldsymbol{I}_6}$ & $\boldsymbol{K}_p^i$ & ${\rm{50}}{\boldsymbol{I}_6}$\\
		
		\bottomrule 
	\end{tabular}
	\vspace{-18pt} 
\end{table}

The point set of the peg position is denoted as ${X_p} = \{ {\boldsymbol{p}_{{\alpha_1}{p_1}}}, \cdots ,{\boldsymbol{p}_{{\alpha_m}{p_m}}}\}$, and the point set of the hole position is denoted as ${X_h} = \{ {\boldsymbol{p}_{{\alpha_1}{h_1}}}, \cdots ,{\boldsymbol{p}_{{\alpha_m}{h_m}}}\}$. The centroids of ${X_p}$ and ${X_h}$ are respectively denoted as:
\begin{equation}
	{ \left \{ \begin{array}{l}
			{\boldsymbol{c}_p} = \frac{1}{m}\sum\limits_{i = 1}^m {{\boldsymbol{p}_{{\alpha_i}{p_i}}}} \\
			{\boldsymbol{c}_h} = \frac{1}{m}\sum\limits_{i = 1}^m {{\boldsymbol{p}_{{\alpha_i}{h_i}}}} 
		\end{array} \right.}
	\label{eq:3}
\end{equation}

Next, we centralize $\boldsymbol{p}_{{\alpha_i}{p_i}}$ and $\boldsymbol{p}_{{\alpha_i}{h_i}}$ and eliminate their components along the ${\boldsymbol{l}}$:
\begin{equation}
	{ \left \{ \begin{array}{l}
			{\boldsymbol{p}'_{{\alpha_i}{p_i}}} = {\boldsymbol{p}_{{\alpha_i}{p_i}}} - {\boldsymbol{c}_p} - \left( {\left( {{\boldsymbol{p}_{{\alpha_i}{p_i}}} - {\boldsymbol{c}_p}} \right) \cdot \boldsymbol{l}} \right)\boldsymbol{l}\\
			{\boldsymbol{p}'_{{\alpha_i}{h_i}}} = {\boldsymbol{p}_{{\alpha_i}{h_i}}} - {\boldsymbol{c}_h} - \left( {\left( {{\boldsymbol{p}_{{\alpha_i}{h_i}}} - {\boldsymbol{c}_h}} \right) \cdot \boldsymbol{l}} \right)\boldsymbol{l}
		\end{array} \right. }
	\label{eq:4}
\end{equation}

The centralized point sets are denoted as ${X'_p} = \{ {\boldsymbol{p}'_{{\alpha_1}{p_1}}}, \cdots ,{\boldsymbol{p}'_{{\alpha_m}{p_m}}}\} $ and ${X'_h} = \{ {\boldsymbol{p}'_{{\alpha_1}{h_1}}}, \cdots ,{\boldsymbol{p}'_{{\alpha_m}{h_m}}}\} $, respectively. The rotation matrix ${\boldsymbol{R}_{ph}}$ between the pegs and holes can be obtained by solving the following least-squares objective function:
\begin{equation}
	{E\left( {{\boldsymbol{R}_{ph}}} \right) = \frac{1}{m}\sum\limits_{i = 1}^m {\left\| {{\boldsymbol{p}'_{{\alpha_i}{h_i}}} - {\boldsymbol{R}_{ph}}{\boldsymbol{p}'_{{\alpha_i}{p_i}}}} \right\|} }
	\label{eq:5}
\end{equation}

The optimal ${\boldsymbol{R}_{ph}}$ can be obtained by using the Umeyama algorithm \cite{c24}. First, we calculate the covariance matrix ${\boldsymbol{H}}$:
\begin{equation}
	{\boldsymbol{H} = {\sum\nolimits_{i = 1}^m {{\boldsymbol{p}'_{{\alpha_i}{p_i}}}\left( {{\boldsymbol{p}'_{{\alpha_i}{h_i}}}} \right)} ^{\rm{T}}} }
	\label{eq:6}
\end{equation}

The SVD decomposition of ${\boldsymbol{H}}$ yields:
\begin{equation}
	{\boldsymbol{H} = \boldsymbol{U}\boldsymbol{\Sigma} {\boldsymbol{V}^{\rm{T}}} }
	\label{eq:7}
\end{equation}

Where $\boldsymbol{\Sigma}$ is a diagonal matrix, $\boldsymbol{U}$ and $\boldsymbol{V}$ are orthogonal matrices. The optimal ${\boldsymbol{R}_{ph}^ *}$ can be obtained:
\begin{equation}
	{{\boldsymbol{R}_{ph}^*} = \boldsymbol{VS}{\boldsymbol{U}^{\rm{T}}} }
	\label{eq:8}
\end{equation}

To ensure det(${\boldsymbol{R}_{ph}^*}$) = 1, the  matrix $\boldsymbol{S}$ is introduced:
\begin{equation}
	{\boldsymbol{S} = \left\{ {\begin{array}{*{20}{c}}
				\boldsymbol{I}_3&{\det (\boldsymbol{V})\det (\boldsymbol{U}) = 1}\\
				{{\rm{diag}}(1,1, - 1)}&{\det (\boldsymbol{V})\det (\boldsymbol{U}) =  - 1}
		\end{array}} \right.}
	\label{eq:9}
\end{equation}

Since the component along the ${\boldsymbol{l}}$ is eliminated in (\ref{eq:4}), we can obtain the orientation error ${d_\theta }$ between the pegs and the holes along the ${\boldsymbol{l}}$:
\begin{equation}
	{{d_\theta } = \arccos \left( {\frac{{tr\left( {{\boldsymbol{R}_{ph}^*}} \right) - 1}}{2}} \right)}
	\label{eq:10}
\end{equation}

The relative position error ${\boldsymbol{d}_p}$ between the pegs and the holes can be directly calculated using the ${\boldsymbol{c}_h}$ and $\boldsymbol{c}_p$:
\begin{equation}
	{{\boldsymbol{d}_p} = {\boldsymbol{c}_h} - {\boldsymbol{c}_p} - \left( {\left( {{\boldsymbol{c}_h} - {\boldsymbol{c}_p}} \right) \cdot \boldsymbol{l}} \right)\boldsymbol{l}}
	\label{eq:11}
\end{equation}

Based on the orientation error ${d_\theta }$ and the position error ${\boldsymbol{d}_p}$, the virtual force ${\boldsymbol{F}}$ and the virtual torque ${\boldsymbol{M}}$ are calculated:
\begin{equation}
	{\left\{ \begin{array}{l}
			\boldsymbol{F} = {k_F}{\boldsymbol{d}_p}/\left\| {{\boldsymbol{d}_p}} \right\|\\
			\boldsymbol{M} = {k_M}{\rm{sign}}\left( {{d_\theta }} \right)
		\end{array} \right.}
	\label{eq:12}
\end{equation}

Then, we divide the assembly process into two stages: before the contact occurs, and after the contact occurs. If the norm of the contact force is greater than $F_c$, contact is said to have occurred. Let ${\boldsymbol{h}_d}$ represent the desired virtual forces and the virtual torque acting on the object, and let ${\rm{S}}_I^i \in \left\{ {{\rm{Obscured, Separated, Overlapped}}} \right\}$ represent the relative state of the peg and hole in the image of the manipulator $i$. We use the relative state to determine when to perform the insertion action. The value of ${\boldsymbol{h}_d}$ is calculated as follows:
\begin{equation}
	\left\{ \begin{array}{l}
		{\boldsymbol{h}_d} = {\left[ {{\boldsymbol{F}^{\rm{T}}} - {k_F}{\boldsymbol{l}^{\rm{T}}},{\boldsymbol{M}^{\rm{T}}}} \right]^{\rm{T}}}{\rm{If\ contact}}\\
		{\boldsymbol{h}_d} = {\left[ { - {k_F}{\boldsymbol{l}^{\rm{T}}},{\boldsymbol{0}_3}} \right]^{\rm{T}}}{\rm{elif\ S}}_I^1{\rm{=}} \cdots {\rm{=S}}_I^m{\rm{=Overlapped}}\\
		{\boldsymbol{h}_d} = {\left[ { - {k_F}{\boldsymbol{l}^{\rm{T}}},{\boldsymbol{0}_3}} \right]^{\rm{T}}}{\rm{elif\ S}}_I^1{\rm{=}} \cdots {\rm{=S}}_I^m{\rm{=Obscured}}\\
		{\boldsymbol{h}_d} = {\left[ {{\boldsymbol{F}^{\rm{T}}},{\boldsymbol{M}^{\rm{T}}}} \right]^{\rm{T}}}{\rm{else}}
	\end{array} \right.
	\label{eq:13}
\end{equation}

Where ${\boldsymbol{0}_3}$ is a zero vector of size $1\times3$. Before contact, if the images from all cameras are in overlapped state or obscured state, the object is inserted along the axis of the hole. Otherwise, the object is moved in the direction that reduces the misalignment. It should be noted that, when the states are all ``Obscured", the ${\boldsymbol{h}_d}$ is related to the grasping pose of the manipulators. In (\ref{eq:13}), it represents the case where the manipulators are opposite to each other as shown in Fig.\ref{fig1}. After contact, while continuing to move in the direction that reduces the misalignment, we also maintain a contact force ${k_F}{\boldsymbol{l}^{\rm{T}}}$ between the pegs and the holes.

\begin{figure*}[t]
	\centering
	\setlength{\abovecaptionskip}{0.cm}
	\includegraphics[width=0.95\textwidth]{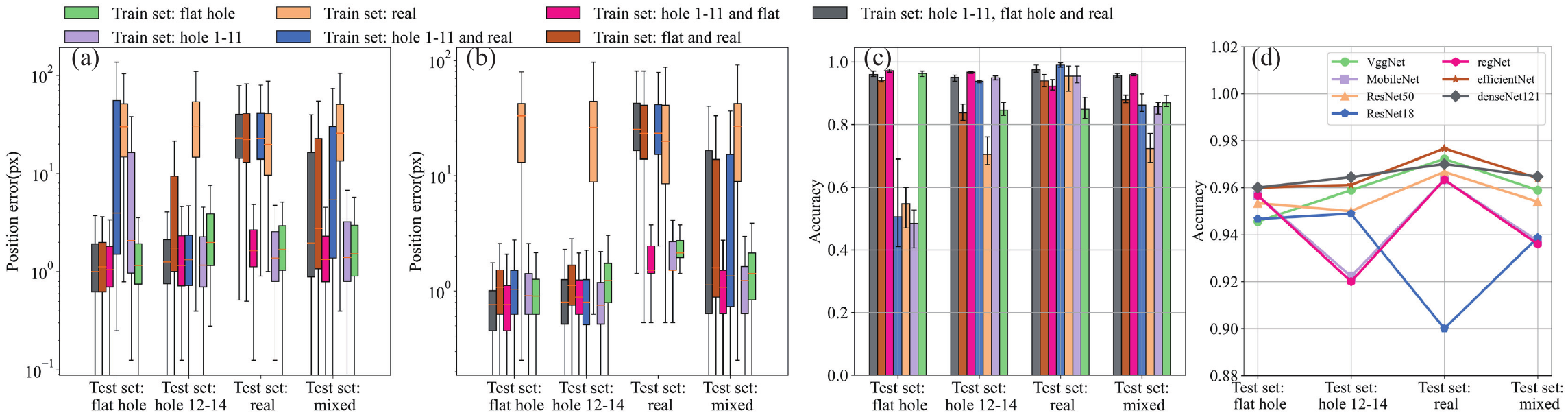} 
	\caption{The test results of the state classification network and the positioning network. (a) The positioning errors of the hole; (b) The positioning errors of the peg; (c) The accuracy of the state classification network (EfficientNetV2-S); (d) Comparison of the accuracy of classical classification networks.}
	\label{fig5}
	\vspace{-10pt} 
\end{figure*}

\begin{table*}[t]
	\scriptsize
	\centering
	\setlength{\abovecaptionskip}{0.cm}
	\caption{The Results of Peg-in-Hole Experiments }
	\setlength{\tabcolsep}{7pt}
	\label{table:2}
	\begin{tabular}{lllllll}
		\toprule
		\textbf{Holes} & \textbf{Initial Deviation} & \textbf{STTM} & \textbf{PBVS} & \textbf{PBVS-SC} & \textbf{PBVS-STTM} & \textbf{VF (ours)} \\
		\midrule
		\multirow{3}{*}{\makecell[l]{Peg1\\and\\Hole1}} 
		& Deviation1 & 35.00$\pm$13.70s (5/15) & 21.76$\pm$10.43s \textbf{(15/15)} & 15.99$\pm$4.46s \textbf{(15/15)} & \textbf{14.41$\pm$5.06s} (8/15) & 14.95$\pm$5.95s \textbf{(15/15)} \\
		& Deviation2 & / & 41.12$\pm$25.77s \textbf{(15/15)} & \textbf{19.95$\pm$3.95s \textbf{(15/15)}} & 32.15$\pm$21.91s (7/15) & 24.10$\pm$4.27s \textbf{(15/15)} \\
		& Deviation3 & / & 52.28$\pm$29.30s (13/15) & 29.36$\pm$3.41s \textbf{(15/15)} & 106.60s (1/15) & \textbf{28.98$\pm$5.41s (15/15)} \\
		\cmidrule{1-7}
		\multirow{3}{*}{\makecell[l]{Peg2\\and\\Hole2}}
		& Deviation1 & 31.33$\pm$13.22s (3/15) & 20.59$\pm$7.31s (13/15) & 21.80$\pm$10.81s (13/15) & 23.96$\pm$31.55s (11/15) & \textbf{18.07$\pm$6.22s (15/15)} \\
		& Deviation2 & / & 36.91$\pm$10.18s (13/15) & 40.58$\pm$10.18s (13/15) & 56.44$\pm$18.87s (3/15) & \textbf{28.67$\pm$3.36s (15/15)} \\
		& Deviation3 & / & 51.31$\pm$21.41s (11/15) & 45.71$\pm$18.44s (13/15) & (0/15) & \textbf{33.58$\pm$5.74s (15/15)} \\
		\bottomrule
	\end{tabular}
	\label{table:result}
	\vspace{-15pt} 
\end{table*}

\subsection{Force Controller for Tightly Coupled Multi-manipulator}

To ensure the safety of the assembly process, it is necessary to control not only the movement of the object under the action of the virtual forces ${\boldsymbol{h}_d}$, but also to keep the internal forces acting on the object within a limited range. To achieve this, we adopt a two-level control scheme. We assume that both the manipulated object and the environment are rigid. We use the object level force control algorithm to generate reference motions for the end effector of the manipulator, and then use an internal force controller at the end effector level to generate joint velocity commands. First, we select a control point $y$ fixed to the object. Then, the reference acceleration of point $y$ is computed using the control law:
\begin{equation}
	{\boldsymbol{\dot v}_{{y_r}}}\boldsymbol{ = }{\left( {\boldsymbol{M}_d^y} \right)^{ - 1}}\left( { - \boldsymbol{K}_d^y{\boldsymbol{v}_y} - \boldsymbol{K}_p^y\left( {{\boldsymbol{h}_d} - {\boldsymbol{h}_c}} \right)} \right)
	\label{eq:14}
\end{equation}

Where ${\boldsymbol{v}_y}{\rm{ = [}}\boldsymbol{\dot p}_y^{\rm{T}},\boldsymbol{\omega }_y^{\rm{T}}{{\rm{]}}^{\rm{T}}}$ represents the velocity of point $y$, ${\boldsymbol{\dot v}_{{y_r}}}{\rm{ = [}}\boldsymbol{\ddot p}_{{y_r}}^{\rm{T}},\boldsymbol{\dot \omega }_{{y_r}}^{\rm{T}}{{\rm{]}}^{\rm{T}}}$ is the reference acceleration of point $y$, $\boldsymbol{M}_d^y \in {{\mathbb{R}}^{6 \times 6}}$, $\boldsymbol{K}_d^y \in {{\mathbb{R}}^{6 \times 6}}$ and $\boldsymbol{K}_p^y \in {{\mathbb{R}}^{6 \times 6}}$, respectively, denote the desired object inertia matrix, damping matrix and force feedback gain coefficient. All three matrices are chosen to be positive symmetric definite matrices, ${\boldsymbol{h}_c}{\rm{ = [}}\boldsymbol{f}_c^{\mathop{\rm T}\nolimits} ,\boldsymbol{\mu }_c^{\mathop{\rm T}\nolimits} {{\rm{]}}^T}$ is the resulting force and the resulting torque applied to the object by the environment. ${\boldsymbol{h}_c}$ can be estimated using force/torque sensors on the end effector. Integrating ${\boldsymbol{\dot v}_{{y_r}}}$ yields the reference motion ${{\cal T}_{{y_r}}}$ of the object, which is composed of the reference position ${\boldsymbol{p}_{{y_r}}}$, the reference rotation matrix ${\boldsymbol{R}_{{y_r}}}$, the reference velocity ${\boldsymbol{v}_{{y_r}}}$, and the reference acceleration ${\boldsymbol{\dot v}_{{y_r}}}$. Assuming there is no relative motion between the end-effector of the manipulator and the object, the desired motion ${{\cal T}_{{i_d}}}$ of the end effector of manipulator $i$ can be calculated by using the closed-chain constraint. The ${{\cal T}_{{i_d}}}$ made up of ${\boldsymbol{p}_{{i_d}}},{\boldsymbol{R}_{{i_d}}},{\boldsymbol{v}_{{i_d}}},{\boldsymbol{\dot v}_{{i_d}}}$, which are the desired position, the desired rotation matrix, the desired velocity and the desired acceleration, respectively. To control the internal forces acting on the object, we use the following end-effector-level impedance controller:
\begin{equation}
	\boldsymbol{M}_d^i\Delta \boldsymbol{\dot v}_{{i_d}i}^{{i_d}} + \boldsymbol{K}_d^i\Delta \boldsymbol{v}_{{i_d}i}^{{i_d}} + \boldsymbol{K}_p^i\Delta {\boldsymbol{x}^{{i_d}}}\boldsymbol{ = h}_{Ii}^{{i_d}}
	\label{eq:15}
\end{equation}

In the equation, $\boldsymbol{M}_d^i \in {{\mathbb{R}}^{6 \times 6}}$, $\boldsymbol{K}_d^i \in {{\mathbb{R}}^{6 \times 6}}$and  $\boldsymbol{K}_p^i \in {{\mathbb{R}}^{6 \times 6}}$ represent the desired inertia matrix, damping matrix and stiffness matrix of the end-effector of the manipulator, respectively, all chosen to be symmetric positive definite matrices. $\Delta {\boldsymbol{x}^{{i_d}}}$ represents the difference between the actual pose and the desired pose of the end-effector of the manipulator $i$. Similarly, $\Delta \boldsymbol{v}_{{i_d}i}^{{i_d}}$ represents the difference between the actual and desired velocity, $\Delta \boldsymbol{\dot v}_{{i_d}i}^{{i_d}}$ represents the difference between the reference acceleration and the desired acceleration. $\boldsymbol{h}_{Ii}^{{i_d}}$ is the component of the force/torque applied by the end-effector to the object that generates internal force/torque within the object. The above variables are expressed in the desired pose coordinate system ${\sum _{{i_d}}}$. Finally, the reference acceleration ${\boldsymbol{\dot v}_{{i_r}}}{\rm{ = [}}\boldsymbol{\ddot p}_{{i_r}}^{\rm{T}}{\rm{,}}\boldsymbol{\dot \omega }_{{i_r}}^{\rm{T}}{{\rm{]}}^{\rm{T}}}$ of the end-effector can be calculated using equation \eqref{eq:15}. The reference velocity ${\boldsymbol{v}_{{i_r}}}$ of the end-effector is obtained by integrating ${\boldsymbol{\dot v}_{{i_r}}}$. Then, the reference joint velocity is calculated as follows:
\begin{equation}
	{\boldsymbol{v}_{{\boldsymbol{q}_i}}} = \boldsymbol{J}{\left( {{\boldsymbol{q}_i}} \right)^\dag }{\boldsymbol{v}_{{i_r}}}
	\label{eq:16}
\end{equation}

Where ${\boldsymbol{q}_i} \in {{\mathbb{R}}^n}$ is the joint angle vector of the $n$-DOF manipulator and $\boldsymbol{J}{\left( {{\boldsymbol{q}_i}} \right)^\dag } \in {{\mathbb{R}}^{n \times 6}}$ represents the pseudoinverse of the Jacobian matrix. ${\boldsymbol{v}_{{\boldsymbol{q}_i}}}$ is the joint velocity command of the manipulator $i$.

\section{EXPERIMENTS}

In this section, we first test the performance of the classification network and the positioning network. Subsequently, the success rate and time consumption of the visual servo control framework were tested in two MPiH tasks.

\subsection{Positioning Accuracy and Classification Accuracy}

\subsubsection{Training Details} The image data used in this paper are divided into three categories, namely, flat hole dataset, shaped hole dataset (holes 1-14), and real dataset. The flat hole dataset contains a total of 1800 images, with 600 images each for obscured, separated, and overlapped states. 1500 images are selected as the training set and the remaining images are used as the test set. The shaped hole dataset contains a total of 4200 images with 14 different hole shapes. There are 300 images for each hole shape, with 100 images each for obscured, separated, and overlapped states. We use the images from holes 1-11 as the training set and images from holes 12-14 as the test set. The real dataset contains 900 images of 3 scenes, and each scene has 300 images (100 images each for obscured, separated, and overlapped states). We use the images of scenes 1-2 as the training set, and the images of scene 3 as the test set. In order to test the effect of different types of images, we combined the three training sets and divided them into seven groups for training, and test were performed on four test sets, as shown in Fig.~\ref{fig5}. Among them, the mixed test set contains flat hole test set, real test set and shaped hole test set (300 images were randomly selected from the images of holes 12-14), totaling 900 images.

\begin{figure}[t]
	\centering
	\setlength{\abovecaptionskip}{0.cm}
	\includegraphics[width=0.45\textwidth]{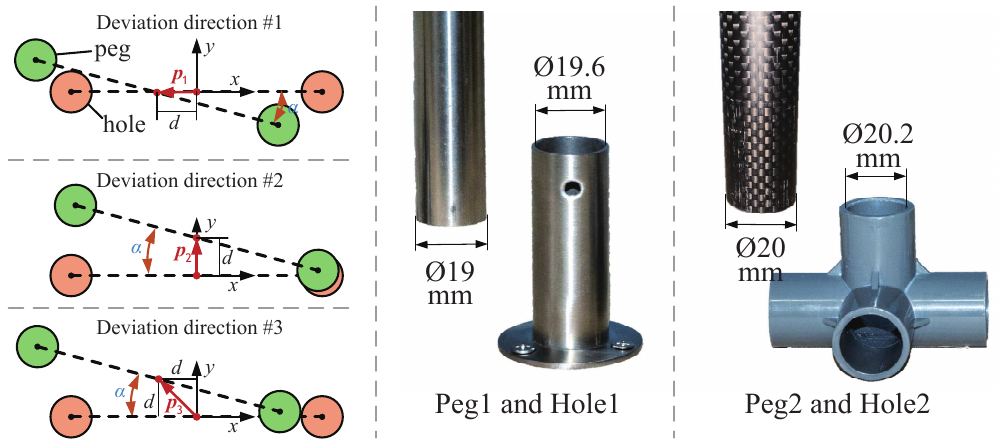} 
	\caption{Left: there directions of the position deviation; Center: the Peg1 and Hole1; Right: the Peg2 and Hole2.}
	\label{fig6}
	\vspace{-15pt} 
\end{figure}

To mitigate the influence of background clutter, similar to \cite{c14}, we randomly select natural images from the MS COCO dataset during training. These natural images are then randomly rotated and overlayed onto the simulated images with a probability of 50\%. Additionally, we introduce ISO noise, Gaussian noise, and Gaussian blur to the images and randomly adjust brightness, contrast, and saturation. We also horizontally flip images with a probability of 50\%. We trained each model for 30 epochs with a batch size of 8, a maximum learning rate of 0.001, a weight decay of 0.0001, and the optimizer is Adam. In addition to the EfficientNetV2-S, we have tested some classic classification networks, including densenet121 \cite{c25}, RegNetY-800MF \cite{c26}, resnet18 \cite{c27}, resnet50 \cite{c27}, MobileNetV3 \cite{c28} and VGG-16 \cite{c29}. Training is carried out on an NVIDIA GeForce RTX 3090. When using the hole 1-11 dataset, it took about 16 minutes for the classification network (EfficientNetV2-S) and about 24 minutes for the positioning network.  

\begin{figure}[t]
	\centering
	\setlength{\abovecaptionskip}{0.cm}
	\includegraphics[width=0.45\textwidth]{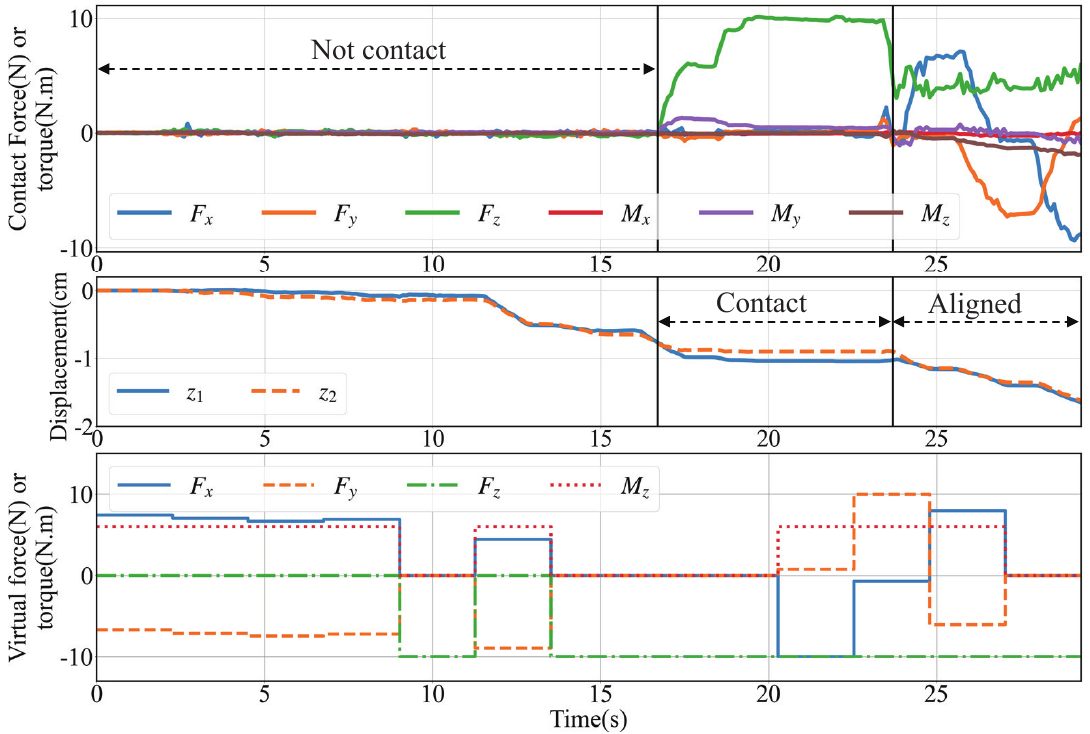} 
	\caption{The state information during the assembly process. Up: the estimated contact force/torque between the peg and the hole. Middle: the displacement of the two end-effectors in the direction of ${\boldsymbol{l}}$. Down: the virtual force/torque ${\boldsymbol{h}_d}$ acting on the object during the assembly process(the ${\boldsymbol{l}}$ coincides with the z-axis of the world coordinate system, the virtual torque is applied only around the z-axis).}
	\label{fig7}
	\vspace{-15pt} 
\end{figure}

\subsubsection{Positioning Accuracy} The distribution of the positioning errors of the pegs and the holes are shown in Fig.~\ref{fig5}(a)-(b). Since the hole may be obscured by the peg, positioning the hole is more challenging than positioning the peg. From the experimental results, it can be seen that training using only flat hole dataset or shaped hole dataset performs well on their corresponding test sets, but the performance decreases on other test sets. Due to the poor diversity of the real data set, training using only the real dataset results in significant overfitting and poor performance on all test sets. Training on synthetic data and real data together also did not improve performance on the real test set.  Among all training schemes, training with data from flat hole dataset and shaped hole dataset together achieves the overall best performance. 

\subsubsection{Classification Accuracy} We first compare the performance of different classification networks. All classification networks are trained with flat hole dataset, shaped hole dataset, and real dataset together. The results are shown in Fig.~\ref{fig5}(d). In the task of this paper, lightweight networks such as efficientNet and denseNet can reach the performance of large networks such as VggNet, the efficientNet is selected for detailed testing in this paper. The efficientNet was trained 3 times on each training set. The average and the upper/lower bounds of accuracy are shown in Fig.~\ref{fig5}(c). According to the experimental results, training with only flat hole dataset or real dataset performs well on their corresponding test sets, but the performance decreases on other test sets. Training with only shaped hole dataset performs well on the real test set and the shaped hole test set. Among all the training schemes, the best overall performance is achieved by training with the flat hole dataset, shaped hole dataset, and real dataset together.

\subsection{Peg-in-hole Performance}

\begin{figure}[t]
	\centering
	\setlength{\abovecaptionskip}{0.cm}
	\includegraphics[width=0.43\textwidth]{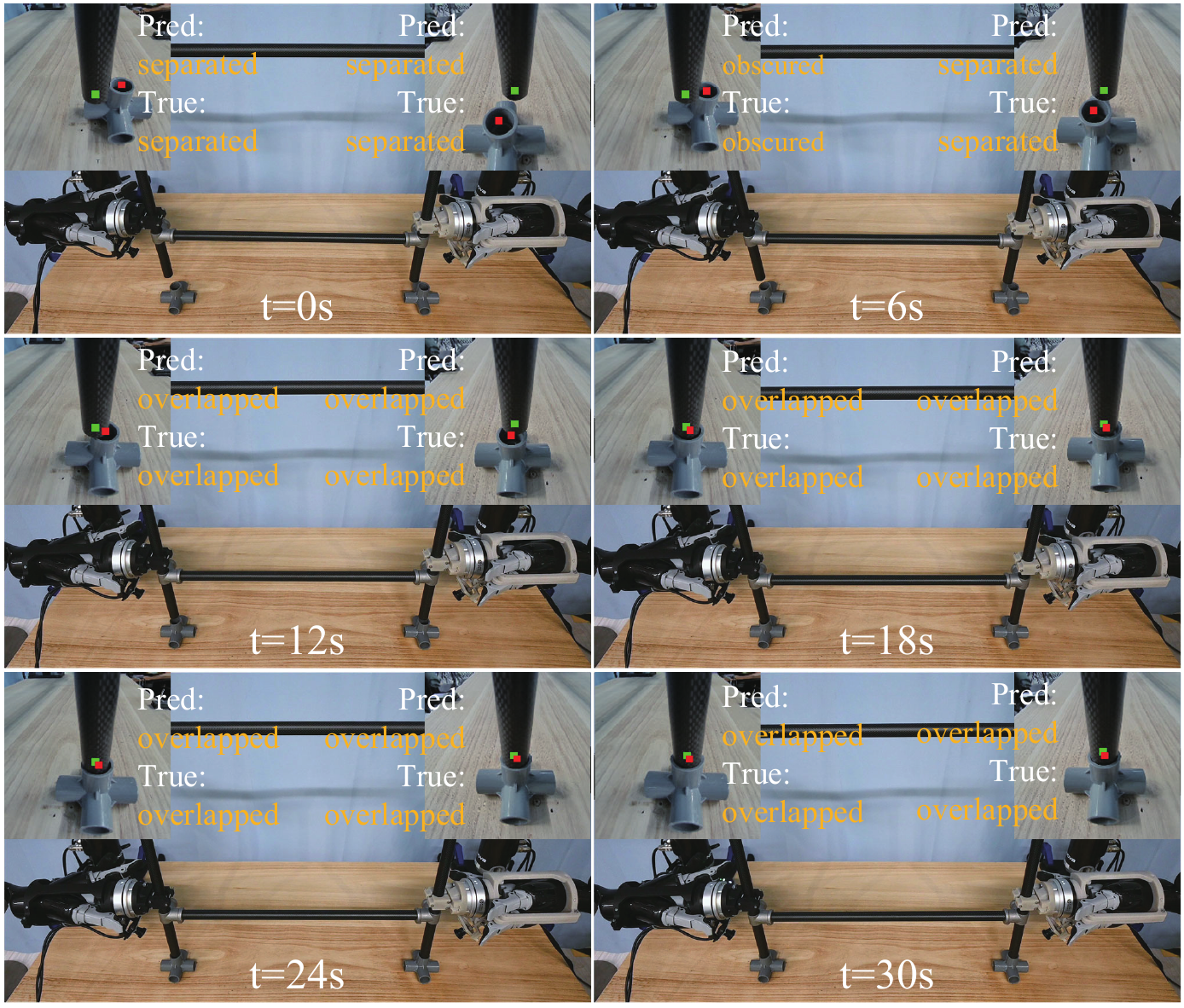} %
	\caption{Some key frames of the assembly process, and the results from the state classification neural network and the positioning neural network.}
	\label{fig8}
	\vspace{-15pt} 
\end{figure}

\subsubsection{Task Settings} We use two Kinova Jaco2 manipulators, as illustrated in Fig.~\ref{fig1}. Each manipulator is equipped with a force/torque sensor and a monocular camera in the hand. We conducted experiments on two MPiH tasks, as shown in Fig.~\ref{fig6}. The distance between the two pegs is 0.5m. In the assembly experiment, the initial deviation between the pegs and the holes is represented by angle $\alpha $ and distance $d$. The direction of position deviation can be divided into three types, as shown in Fig.~\ref{fig6}. We set three different deviation sizes (\textbf{Deviation1:} $\alpha  = 1.5^\circ $, $d$ = 0.5cm; \textbf{Deviation2:} $\alpha  = 3^\circ $, $d$ = 1.0cm; \textbf{Deviation3:} $\alpha  = 4.5^\circ $, $d$ = 1.5cm), and 15 assembly experiments were performed under each deviation value (5 experiments were performed in each deviation direction). In the case of Deviation1, the end faces of the pegs and the holes are in contact, and in the case of Deviation 2 and 3, the end faces of the pegs and the holes are 1cm apart. The other parameters are shown in TABLE~\ref{table:1}. A laptop computer with GeForce GTX 1050 Ti GPU and Inter i7-8750H CPU was used for the assembly experiments. Due to the limited computing power of the computer, the virtual force is re-calculated every 1.3s. Each calculation of the virtual force takes about 0.7s, during which the manipulator will stop moving. If all the pegs and holes are aligned, the task is considered successful. If the assembly could not be completed in 150s or the pegs slips
outside the holes during the search, the assembly was considered failure.

\subsubsection{Baseline Algorithms} We compared our proposed method with the following baselines: 

\textbf{Baseline 1: STTM.} STTM \cite{c19} is a variant of the spiral trajectory search for the MPiH task. The trajectory of STTM can be described by four parameters $p$, ${v_{{\rm{spiral}}}}$, ${\omega _{{\rm{spiral}}}}$, and ${\theta _{{\rm{spiral}}}}$, $p$ is the spiral pitch, ${v_{{\rm{spiral}}}}$ is the linear velocity along the spiral trajectory, ${\omega _{{\rm{spiral}}}}$ is the angular velocity along the $\boldsymbol{l}$, ${\theta _{{\rm{spiral}}}}$ is the search range along the $\boldsymbol{l}$. In this paper, the values of $p$, ${v_{{\rm{spiral}}}}$, ${\omega _{{\rm{spiral}}}}$, and ${\theta _{{\rm{spiral}}}}$ are 0.15 mm, 3.5 mm/s, 0.14 rad/s, and 0.045 rad, respectively. Since the pegs and the holes cannot contact when the initial deviation is large, the STTM is only tested in the case of Deviation1.

\textbf{Baseline 2: Pose-Based Visual Servo (PBVS).} To demonstrate the advantages of the virtual-force based visual servo, we introduced the conventional PBVS \cite{c30} into the MPiH task as the baseline algorithm. First, we calculate the position error  $\boldsymbol{d}_{p}$ and orientation error $d_{\theta}$ based on the positioning results. Then, the reference linear velocity ${\boldsymbol{v}_r}$ and reference angular velocity ${\omega_r}$ of the pegs is calculated:

\begin{equation}
	{\left\{ \begin{array}{l}
			{\boldsymbol{v}_r} = {v_p}{\boldsymbol{d}_{p}}/\left\| {{\boldsymbol{d}_{p}}} \right\|\\
			{\omega _r} = {\omega _p}{\rm{sign}}\left( {{d_{\theta}}} \right)
		\end{array} \right.}
	\label{eq:17}
\end{equation}

Where ${v_p}$ and ${\omega_p}$ represent the size of linear velocity and angular velocity, respectively. Then, the pegs are controlled to move towards the holes at the reference velocity. When both the position error is less than the position threshold ${\varepsilon _p}$ and the orientation error is less than the orientation threshold ${\varepsilon _\theta }$, the pegs are made to perform an insertion movement along the direction of $\boldsymbol{l}$. After completing an insertion movement, if all the pegs and holes are aligned, the task is considered successful. If any peg slips out of the hole, the task is considered a failure. If the above termination conditions do not occur within the time limit, the $\boldsymbol{d}_{p}$ and $d_{\theta}$ are recalculated and the above process is repeated. In this paper, the values of ${v_p}$, ${\omega_p}$, ${\varepsilon _p}$, and ${\varepsilon _\theta }$ are set to 0.02 m/s, 0.15 rad/s, 0.1 mm, and 0.002 rad respectively.

\textbf{Baseline 3: PBVS with State Classification (PBVS-SC).} In the PBVS, the insertion movement will be performed only when the pegs reach the target pose. In the PBVS-SC, we utilize the relative states to determine when to perform the insertion movement, which is similar to the VF method. Before the contact occurs, if the states are all "Obscured" or all "Overlapped", we perform the insertion movement. 

\textbf{Baseline 4: PBVS-STTM.} The magnitude of the initial deviation between the pegs and the holes is an important factor affecting the time consumption of the STTM. In this paper, we combine PBVS and STTM as a baseline algorithm. Firstly, we calculate the initial position error and initial orientation error and try to eliminate this initial error using PBVS. Then, we use the STTM algorithm to search.

\begin{figure}[t]
	\centering
	\setlength{\abovecaptionskip}{0.cm}
	\includegraphics[width=0.46\textwidth]{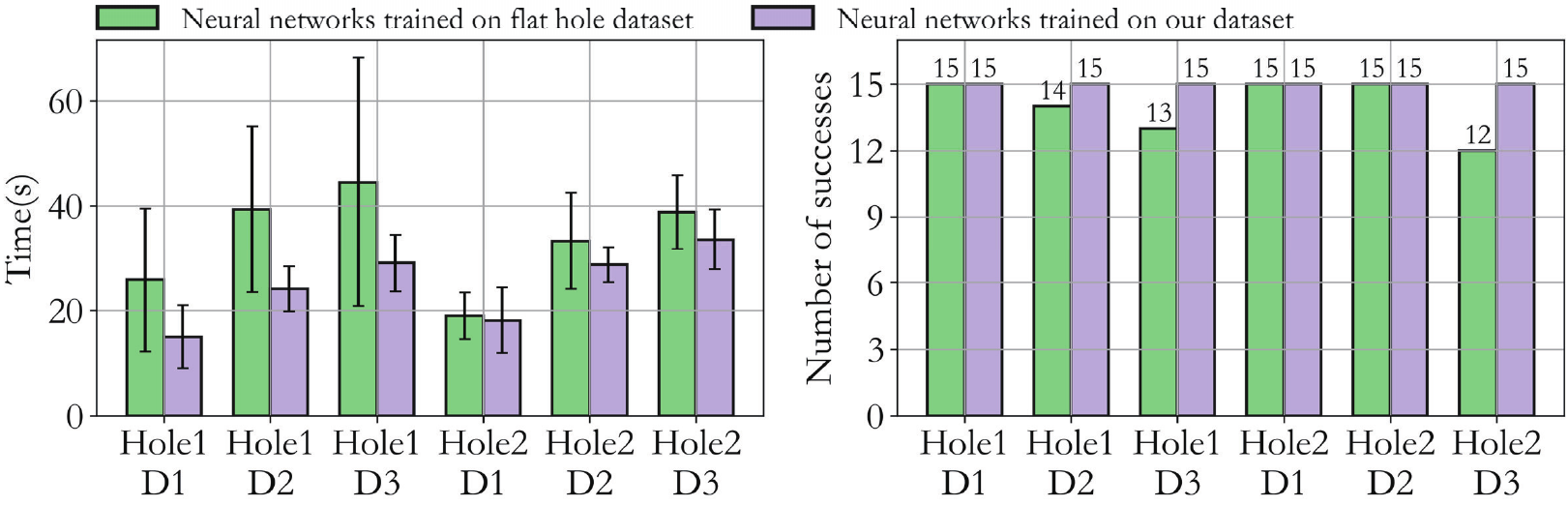} 
	\caption{MPiH performance of neural networks trained with different datasets, Di is the abbreviation of Deviation i. Left: mean and standard deviation of assembly time. Right: the number of successful assemblies.}
	\label{fig9}
	\vspace{-15pt} 
\end{figure}

\subsubsection{The Performance of Different Algorithms} We record the success rate and the time consumption for the experiments, as shown in TABLE~\ref{table:2}. The experimental results show that the proposed VF-based visual servo method achieves the best performance overall. Specifically, among all algorithms, STTM performs the worst, which is due to the fact that it requires a lot of search time in the absence of visual information, and when one of the peg is aligned with the hole, it is easy to jam, and the rest of pegs and the holes cannot be aligned. Although the PBVS can be used to eliminate most of the initial deviation, the above problems still exist. The PBVS and PBVS-SC perform significantly better than the STTM algorithm, especially when the clearance between the peg and hole is large (Peg1 and Hole1). In the PBVS-SC algorithm, benefit from the relative states, the pose error $\boldsymbol{d}_{p}$ and $d_{\theta}$ becomes smaller when the pegs and holes come into contact. When the clearance is large, the PBVS-SC can complete the task more quickly. However, when the clearance is small (Peg2 and Hole2), the effect is not obvious. The reason for the above phenomenon is that the insertion process in PBVS and PBVS-SC is intermittent. When the clearance is small, the peg can easily miss the hole. In the VF, a contact force will always be maintained between the pegs and the holes, and when the peg is aligned with the hole, the peg will slide into the hole, which have higher efficiency.

We selected one successful assembly process from the experiments of Hole2 in the case of Deviation3. Fig.~\ref{fig7}. shows the estimated contact force ${\boldsymbol{h}_c}$ between the pegs and the holes, the displacement of the two end-effectors in the direction of ${\boldsymbol{l}}$ and the virtual force ${\boldsymbol{h}_d}$ acting on the pegs. The key frames and the output of the state classification network and the positioning network are shown in Fig.~\ref{fig8}. 

\subsubsection{The Influence of Dataset} We contrasted the MPiH performance of neural networks trained with different datasets. Specifically, there are two cases. In the first case, both the positioning network and the classification network are trained using only the flat hole dataset. In the second case, the positioning neural network is trained with the flat hole dataset and the shape hole dataset, while the classification network is trained with the flat hole dataset, the shape hole dataset, and the real dataset. The results are shown in Fig.~\ref{fig9}. Thanks to the accurate positioning and state classification, compared with training with flat hole dataset only, using our proposed dataset can achieve higher success rate, while using less time.

\begin{figure}[t]
	\centering
	\setlength{\abovecaptionskip}{0.cm}
	\includegraphics[width=0.46\textwidth]{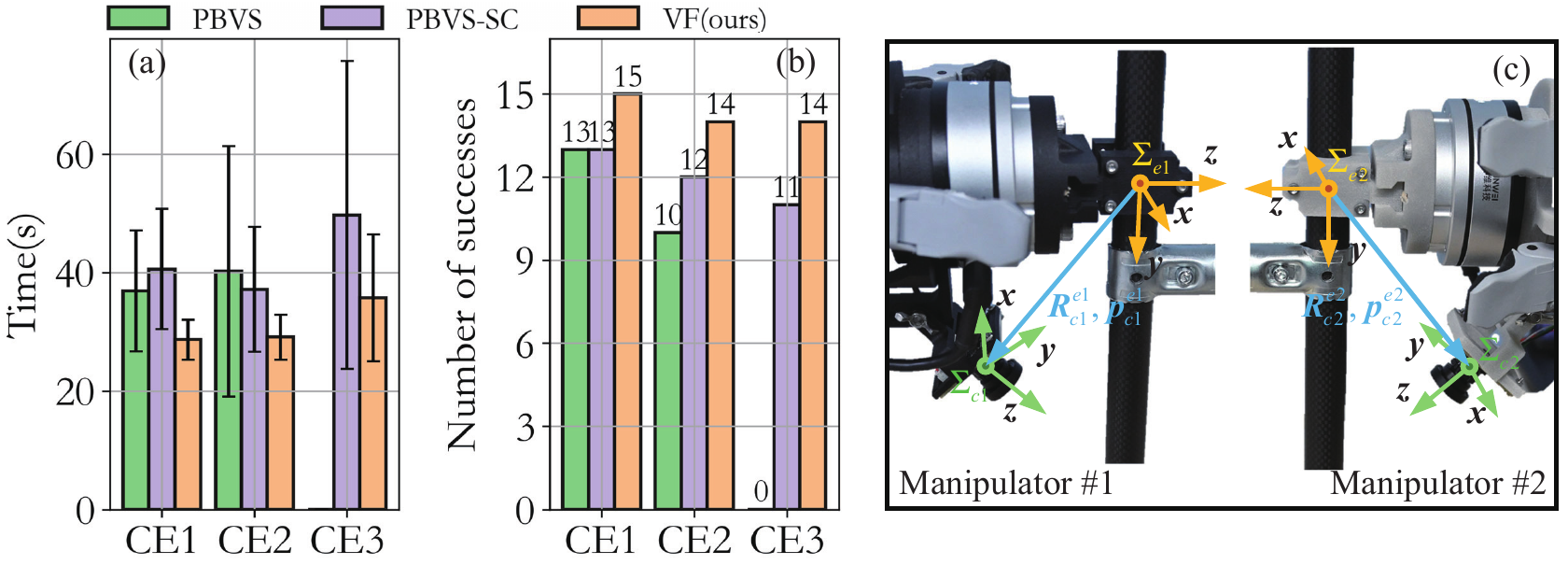} 
	\caption{MPiH performance under different camera calibration errors. (a) Mean and standard deviation of assembly time. (b) The number of successful assemblies. (c) The position of the camera coordinate system and the end-effector coordinate system.}
	\label{fig10}
	\vspace{-15pt} 
\end{figure}

\subsubsection{The Influence of Camera Calibration Errors} 
\begin{figure}[t]
	\centering
	\setlength{\abovecaptionskip}{0.cm}
	\includegraphics[width=0.42\textwidth]{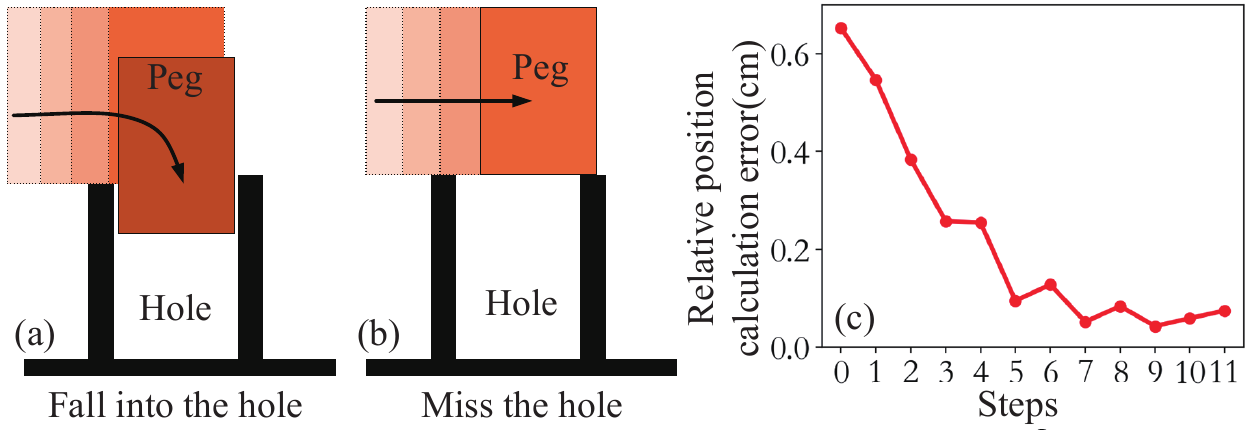} 
	\caption{The diagram of the continuous insertion and the intermittent insertion, and the calculation error of the relative position. (a)The continuous insertion in the VF method, which is achieved by maintaining the contact force between the pegs and holes. (b) The intermittent insertion in the PBVS-SC and PBVS methods. (c) The calculation error of the relative position between the pegs and holes during assembly (in the case of CE2).}
	\label{fig11}
	\vspace{-15pt} 
\end{figure}

Camera calibration error is a key factor affecting visual servoing. The positions of the camera coordinate system ${\Sigma _{ci}}$ and the end-effector coordinate system ${\Sigma _{ei}}$ of manipulator $i$ are shown in Fig.~\ref{fig10}(c). The position and rotation matrices of the ${\Sigma _{ci}}$ with respect to the ${\Sigma _{ei}}$ are denoted as $\boldsymbol{p}_{ci}^{ei}$ and $\boldsymbol{R}_{ci}^{ei}$, respectively. We do experiments by manually adding attitude error ${\boldsymbol{R}_{error}}$ and position error ${\boldsymbol{p}_{error}}$ on the basis of $\boldsymbol{R}_{ci}^{ei}$ and $\boldsymbol{p}_{ci}^{ei}$, and compare three kinds of calibration errors. \textbf{CE1:} Directly use $\boldsymbol{R}_{ci}^{ei}$ and $\boldsymbol{p}_{ci}^{ei}$ obtained by careful calibration, without manually adding attitude error and position error; \textbf{CE2:} The PRY Euler angle corresponding to ${\boldsymbol{R}_{error}}$ is $[10^\circ ,10^\circ ,10^\circ ]$, and the value of the position error ${\boldsymbol{p}_{error}}$ is [0.5,0.5,0.5]cm. \textbf{CE3:} The PRY Euler angle corresponding to ${\boldsymbol{R}_{error}}$ is $[20^\circ ,20^\circ ,20^\circ ]$, and the value of the position error ${\boldsymbol{p}_{error}}$ is [1,1,1]cm. We did experiments in the case of Hole2 Deviation2 and 5 experiments were performed in each deviation direction. The results are shown in Fig.~\ref{fig10}. It can be seen that compared with the PBVS, the VF and PBVS-SC have better robustness. The main reason is that the insertion movement in the PBVS will be performed only when the pegs reach the target pose. If the target pose has a large error (in the case of CE2), the pegs will fall out of the holes and cause the task to fail. In the VF and PBVS-SC, we use the relative states between the pegs and holes to more accurately determine when the insertion movement should be performed. The relative pose $\boldsymbol{d}_{p}$ and $d_{\theta}$ are only used to calculate the movement direction of the pegs. The relative states cannot prevent camera calibration errors from affecting the relative pose, but as the peg gets closer to the hole, the influence of the camera calibration error on the relative pose becomes smaller and smaller. This makes the VF and PBVS-SC methods to maintain a high success rate. Take the assembly process in Fig.~\ref{fig7} as an example, the relative position calculation error caused by the camera calibration error in the case of CE2 is shown in Fig.~\ref{fig11}(c). In the PBVS-SC method, the insertion is intermittent, even a small relative pose calculation error may cause the peg to fail to align with the hole, as shown in Fig.~\ref{fig11}(a)-(b). Therefore, the performance of the VF method is better than that of the PBVS-SC method.

\section{CONCLUSION}

In the MPiH tasks for large-sized parts, contact force cannot provide guidance for hole search. In this paper, we proposes a new synthetic dataset to improve the performance  of the neural network model for extracting visual features. At the same time, we propose the concept of virtual force to integrate the visual features of different positions, it can be naturally combined with the collaborative force controller of multiple manipulators without precisely calibrated camera and additional motion planning. Our method achieves a success rate close to 100\% in the submillimeter-level dual-manipulator dual-hole assembly task. At the same time, it shows strong robustness against the calibration errors of the cameras. This paper assumes that the end-effectors are fixed to the pegs. How to use multiple dexterous hands to complete the MPiH assembly task, and how to achieve the human-robot collaborative assembly are the future research directions. 

\ifCLASSOPTIONcaptionsoff
  \newpage
\fi



%




\end{document}